\documentclass[sigconf]{acmart}

\usepackage{multirow}  
\usepackage{subcaption}
\renewcommand\footnotetextcopyrightpermission[1]{} 
\setcopyright{none}
\AtBeginDocument{%
  }

\begin{document}


\title{Towards Open-world Generalized Deepfake Detection: General Feature Extraction via Unsupervised Domain Adaptation}

\author{Midou Guo}
\email{guomd5@mail2.sysu.edu.cn}
\orcid{0009-0002-0821-986X}
\affiliation{%
  \institution{School of Computer Science and Engineering \\Sun Yat-sen University}
  \city{Guangzhou}
  \country{China}
}

\author{Qilin Yin}
\email{yinqlin@mail2.sysu.edu.cn}
\orcid{0000-0001-7571-046X}
\affiliation{%
  \institution{School of Computer Science and Engineering \\Sun Yat-sen University}
  \city{Guangzhou}
  \country{China}
}

\author{Wei Lu}
\orcid{0000-0002-4068-1766}
\authornote{Corresponding author.}
\email{luwei3@mail.sysu.edu.cn}
\affiliation{
  \institution{School of Computer Science and Engineering \\Sun Yat-sen University}
  \city{Guangzhou}
  \country{China}
  }

\author{Xiangyang Luo}
\orcid{0000-0003-3225-4649}
\email{luoxy\_ieu@sina.com}
\affiliation{
  \institution{State Key Laboratory of Mathematical Engineering \\and Advanced Computing}
  \city{Zhengzhou}
  \country{China}
} 




\renewcommand{\shortauthors}{Midou Guo et al.}

\begin{abstract}
With the development of generative artificial intelligence, new forgery methods are rapidly emerging. Social platforms are flooded with vast amounts of unlabeled synthetic data and authentic data, making it increasingly challenging to distinguish real from fake. Due to the lack of labels, existing supervised detection methods struggle to effectively address the detection of unknown deepfake methods. Moreover, in open world scenarios, the amount of unlabeled data greatly exceeds that of labeled data. Therefore, we define a new deepfake detection generalization task which focuses on how to achieve efficient detection of large amounts of unlabeled data based on limited labeled data to simulate a open world scenario. To solve the above mentioned task, we propose a novel Open-World Deepfake Detection Generalization Enhancement Training Strategy (OWG-DS) to improve the generalization ability of existing methods. Our approach aims to transfer deepfake detection knowledge from a small amount of labeled source domain data to large-scale unlabeled target domain data. Specifically, we introduce the Domain Distance Optimization (DDO) module to align different domain features by optimizing both inter-domain and intra-domain distances. Additionally, the Similarity-based Class Boundary Separation (SCBS) module is used to enhance the aggregation of similar samples to ensure clearer class boundaries, while an adversarial training mechanism is adopted to learn the domain-invariant features. Extensive experiments show that the proposed deepfake detection generalization enhancement training strategy excels in cross-method and cross-dataset scenarios, improving the model's generalization. 

\end{abstract}





\maketitle

\section{Introduction}

\begin{figure*}[t]
\centering
\begin{minipage}{0.3\textwidth} 
    \centering
    \includegraphics[width=1.0\linewidth]{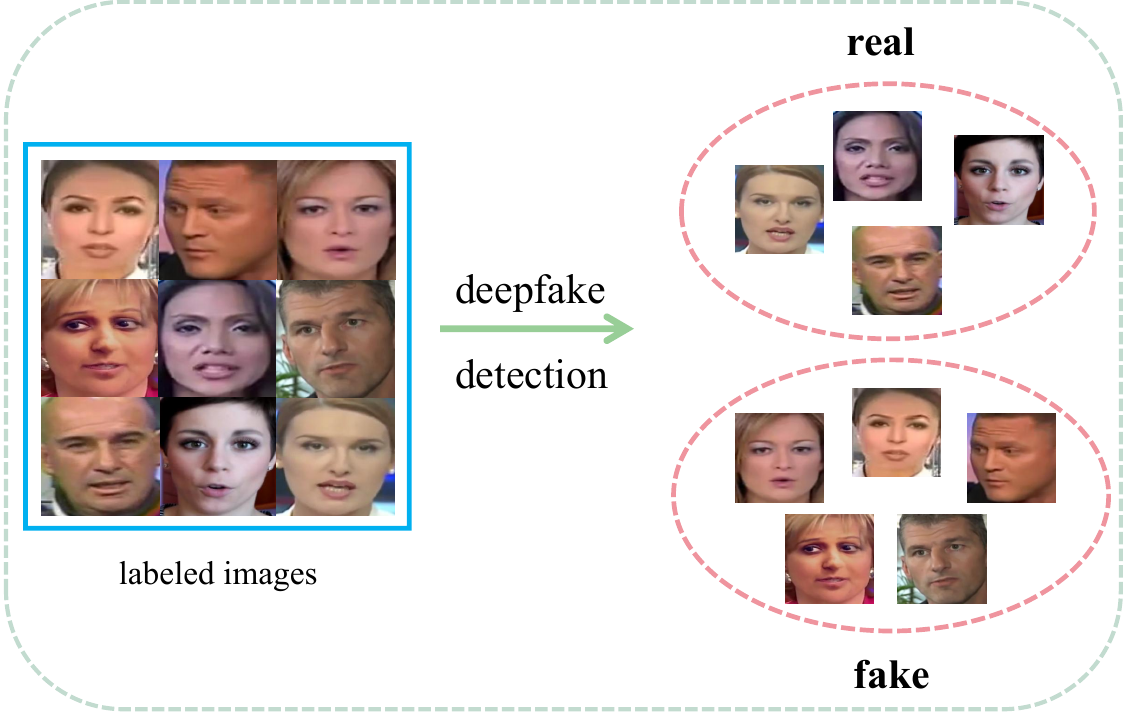} 
    \subcaption{\footnotesize  Traditional deepfake detection}
    \label{fig:a}
\end{minipage}%
\hspace{0.05cm} 
\begin{minipage}{0.3\textwidth} 
    \centering
    \includegraphics[width=1.0\linewidth]{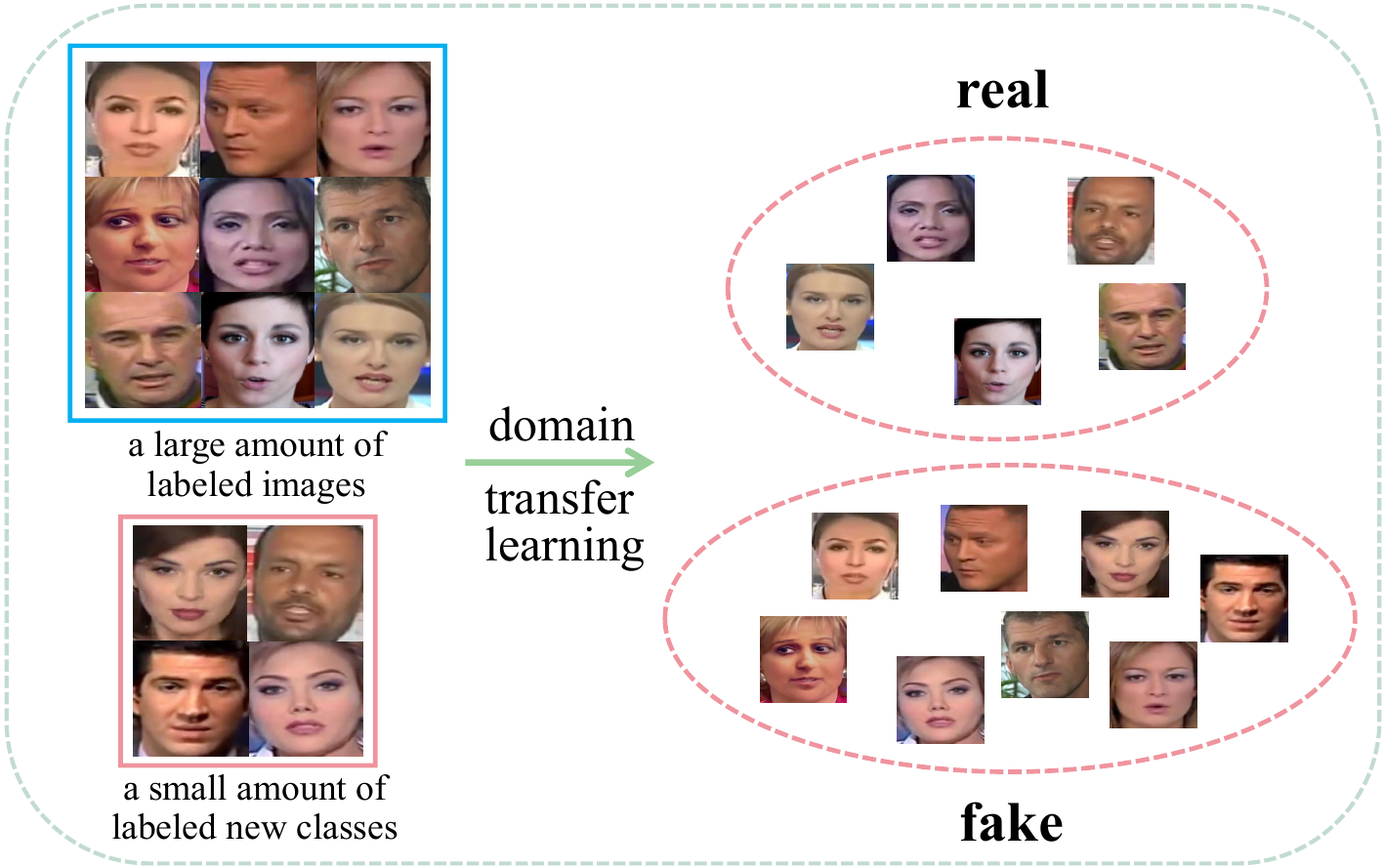} 
    \subcaption{\footnotesize  Traditional domain transfer learning}
    \label{fig:b}
\end{minipage}
\hspace{0.05cm}
\begin{minipage}{0.3\textwidth} 
    \centering
    \includegraphics[width=1.0\linewidth]{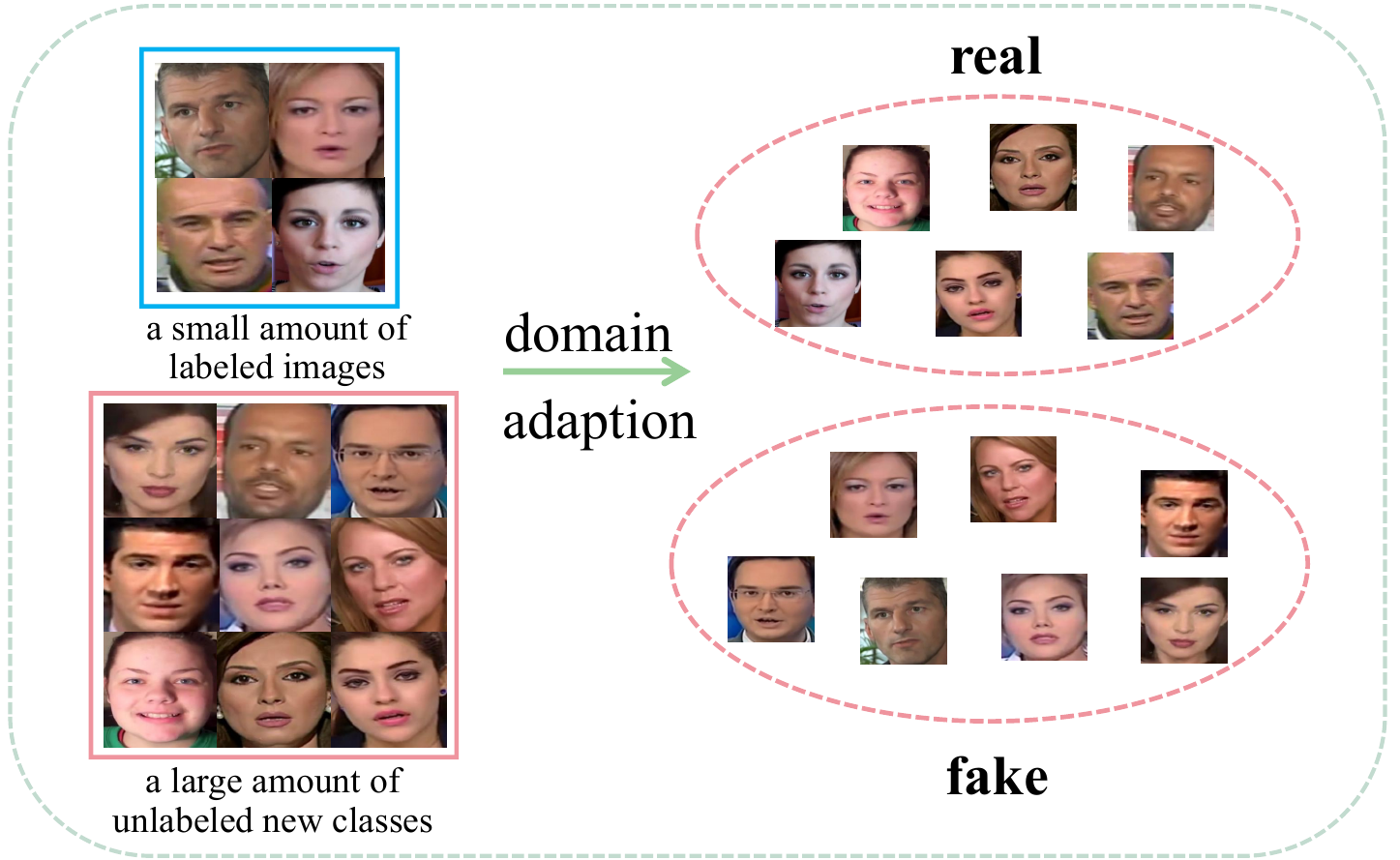} 
    \subcaption{\footnotesize  Open-world unsupervised deepfake detection}
    \label{fig:c}
\end{minipage}

\caption{Illustration of different deepfake detection tasks: (a) Traditional detection relies on labeled data to classify data as "real" or "fake".  (b) Traditional domain transfer learning leverages a mix of large-scale labeled data and a small batch of labeled new forgeries to adapt to unseen manipulations. (c) Open-world deepfake detection simulates open world scenarios where new labels are scarce, utilizing limited labeled data to effectively
detection extensive unlabeled forgeries.}
\label{fig:abc}
\end{figure*}

With the development of generative artificial intelligence technologies, the effectiveness of deepfake generation techniques has significantly improved. These technologies make synthetic facial data more realistic. The widespread use of these forgery techniques has also brought risks of abuse, posing a serious threat to social stability and personal privacy. Therefore, it is urgent and crucial to conduct in-depth research on deepfake detection technologies.

In recent years, deepfake detection methods have made significant progress \cite{li2018ictu,afchar2018mesonet,li2021frequency,sun2021improving,he2021forgerynet,wang2023altfreezing,xu2023tall}, achieving high performance on datasets like FaceForensics++ \cite{rossler2019faceforensics++}. Li et al. \cite{li2020face} proposed "Face X-ray" for detecting forgeries by reconstructing boundaries between forged and original areas, which is effective for traditional GAN-based forgeries. However, this approach is limited when applied to high-quality 
new generation techniques, such as diffusion models \cite{ho2020denoising}. It is worth noting that the high detection performance of these methods was achieved when the training and testing datasets were in the same distribution. However, with the development of generative artificial intelligence technologies, new types of forged samples continue to emerge, and their data distribution differs greatly from known types of forgery. In the absence of effective labels, the above method is difficult to adapt to new data distributions through retraining. That is to say, these methods have poor detection generalization and cannot effectively detect new types of forged samples. 

Recently, several methods improve generalization ability through data augmentation \cite{li2018exposing,zhao2021learning,shiohara2022detecting,chen2022self,wang2022lisiam,wang2023dire} and transfer learning \cite{cozzolino2018forensictransfer,aneja2020generalized,qiu2023few}. Data augmentation methods enhance generalization by simulating forgery variations but this approach relies on large amounts of labeled data. Transfer learning typically enhances generalization by transferring knowledge from the source domain to the target domain. Aneja et al. proposed a meta learning based framework for few-shot/zero-shot facial forgery detection by integrating boundary inconsistencies with global semantic features, requiring large scale labeled data for training to establish robust cross domain adaptation. Qiu et al. \cite{qiu2023few} introduced guided adversarial interpolation to generate discriminative intermediate samples between real and fake images, coupled with adaptive contrastive loss to detect feature mutations, while still dependent on annotated data for training set construction. While transfer learning can effectively improve the generalization of detection methods, it still requires a large amount of labeled data for training, yet obtaining large amounts of labeled data in the open world is difficult.

In the open world, labeled data is often scarce due to the time and effort required for annotation, while unlabeled data can be easily obtained in large quantities. Therefore, we argue that how to efficiently detect large amount of unlabeled data using a small amount of labeled data is a new deepfake detection generalization task. The task is suitable for the detection needs of open world scenarios. As shown in Figure \ref{fig:abc}, our task differs from traditional deepfake detection and domain transfer learning. Traditional deepfake detection task relies heavily on extensive labeled datasets to accurately differentiate between authentic and synthetic data. Traditional domain transfer learning addresses distribution divergence between source and target domains by transferring knowledge from a large amount of labeled source domain data to improve the generalization of limited labeled target data. Both traditional deepfake detection and transfer learning tasks rely heavily on large-scale labeled data. However, our task differs from the above two traditional tasks in that it aims to simulate a deepfake detection task in a complex open-world scenarios, utilizing limited labeled data to effectively detection extensive unlabeled data. 


To address the above open-world deepfake detection generalization task, we propose a unsupervised training strategy termed OWG-DS. In general, there are significant feature distribution differences between labeled and unlabeled data, i.e., domain shift, leading to poor generalization of existing detection methods. Therefore, the OWG-DS proposes to reduces domain shift by aligning the distribution of different domain features to achieve enhanced generalization of existing detection methods.

Specifically, we propose a Domain Distance Optimization (DDO) module, which aligns different domain features by reducing the distance between different domains and increasing intra-domain distances. To mitigate category boundary blurring during feature alignment, we propose a Similarity-based Class Boundary Separation (SCBS) module. By focusing on neighboring samples based on learned similarity metrics, this module helps reinforce the cohesion of the same class samples while pushing apart samples from different classes, thereby enhancing the distinctness of class boundaries and minimizing overlap between classes. To further enhance different domain feature alignment, we introduce an Adversarial Domain Classifier (ADC)  to bridge the feature distributions of different domains and facilitate the convergence between different domains.
Through the collaborative optimization of these modules, the OWG-DS maximizes feature alignment and significantly mitigates domain shift, thereby enhancing the generalization of detection in open world scenarios.


The main contributions of this work are as follows:
\begin{itemize}
    \item We propose a new deepfake detection generalization task for simulating deepfake detection task within open-world scenarios, focusing on how to achieve efficient detection of extensive unlabeled data based on limited labeled data.
    \item We propose a novel Open-World Deepfake Detection Generalization Enhancement Training Strategy (OWG-DS). This strategy aims to transfer deepfake detection knowledge from a small amount of labeled source domain data to large-scale unlabeled target domain data ,thereby enhancing the generalization capability of detection methods.
    \item We design the DDO and SCBS modules to achieve robust feature representation learning through co-optimization, effectively aligning feature distributions and extracting shared features across different domains.

\end{itemize}

\section{Related Work}

Domain Adaptation (DA) aims to transfer knowledge from a known source domain to an unlabeled target domain, thereby enhancing the generalization of detection methods. Based on the availability and quality of target domain labels, domain adaptation can be categorized into Supervised DA (SDA), Weakly-Supervised DA (WSDA), and Unsupervised DA (UDA), with increasing levels of difficulty.


Supervised Domain Adaptation (SDA) utilizes labeled data in both source and target domains for joint optimization to enhance model generalization. For example, Motiian et al. \cite{motiian2017unified} proposed semantic alignment loss and separation loss, ensuring that same-class samples across domains are brought closer, while different-class samples remain distinct. Xu et al. \cite{xu2024identity} proposed an identity-based multimedia forgery detection approach that uses identity information (e.g., facial features) and multimodal data to enhance the accuracy and robustness of forgery detection.

Weakly-supervised domain adaptation (WSDA) assumes that the target domain data is only partially labeled (e.g., 
only image-level labels without pixel-level annotations). For instance, Laine and Aila \cite{laine2016temporal} proposed Temporal Ensembling, which generates stable pseudo-labels by averaging predictions over multiple time steps; Cao et al. \cite{cao2021open} employed a supervised loss for labeled data and a pairwise loss for unlabeled data, incorporating an uncertainty-based adaptive margin to enhance discriminative power and help bridge the gap between known and new classes during training. Miyato et al. \cite{miyato2018virtual} introduced VAT (Virtual Adversarial Training) to improve model robustness through small perturbations; Lee et al. \cite{lee2013pseudo} proposed Self-training, which optimizes training using high-confidence samples; and Zou et al. \cite{zou2018unsupervised} developed Class-Balanced Self-Training, which dynamically adjusts class distributions to improve the recognition of low-frequency categories.

Unsupervised domain adaptation (UDA) trains a model using labeled source domain data and unlabeled target domain data, aiming to achieve strong performance in the target domain. For example, Tzeng et al. \cite{tzeng2017adversarial} introduced ADDA (Adversarial Discriminative Domain Adaptation), which aligns feature distributions across domains using adversarial learning; Ganin and Lempitsky \cite{ganin2015unsupervised} proposed DANN (Domain Adversarial Neural Network), which reduces domain gaps via a gradient reversal layer; Liu et al. \cite{liu2019separate} used a coarse to fine weighting mechanism, the target domain samples are gradually divided into known and unknown categories, and their importance is simultaneously considered in the feature distribution alignment process; and Lv et al. \cite{lv2024domainforensics} introduced the DomainForensics framework, employing a bidirectional domain adaptation mechanism combined with self-distillation to extract useful knowledge from unlabeled target domain data. Zhou et al. \cite{zhou2024fine} applies Network Memory-based Adaptive Clustering (NMAC) to cluster the unlabeled data and uses Pseudo-Label Generation (PLG) to match unlabeled data from the target domain with known forgery methods from the source domain  and generate pseudo-labels. The model is jointly trained with labeled data from the source domain and pseudo-labeled data from the target domain to enhance its ability to detect new forgery methods.

Although existing domain adaptation methods have achieved some success in feature alignment between the source and target domains, these methods typically require a large amount of labeled data to retrain the model for identifying unseen data. However, obtaining a large amount of labeled data in an open world environment is both time-consuming. Therefore, we propose a novel Open-World Deepfake Detection Generalization Enhancement Training Strategy (OWG-DS) to achieve generalized detection of a large amount of unlabeled new fake data using limited labeled data.

\section{Proposed Method}
\subsection{Task Definition}
The open-world deepfake detection generalization task focuses on efficiently detecting large amounts of unlabeled data using limited labeled data. The task involves a labeled source domain dataset $D_S = {(x_i, y_i)}_{i=1}^n$ and an unlabeled target domain dataset $D_T = {(x_i)}_{i=1}^m$, where $x_i$ represents image data and $y_i$ is the label (real or fake), $n$ and $m$ respectively represent the number of samples in the source domain dataset and the target domain dataset. It is worth noting that $n \ll m$. Additionally, due to the characteristics of generalization detection in open world scenarios, this task requires that the forged datasets of the source and target domains be strictly mutually exclusive. Formally, 
\begin{equation}
    D_{\text{source}}^{\text{fake}} \cap D_{\text{target}}^{\text{fake}} = \emptyset,
\end{equation}
where $D_{\text{source}}^{\text{fake}}$ and $D_{\text{target}}^{\text{fake}}$ respectively represent the fake data in the source domain and the fake data in the target domain. 

\begin{figure*}[h] 
    \centering
    \includegraphics[width=\textwidth]{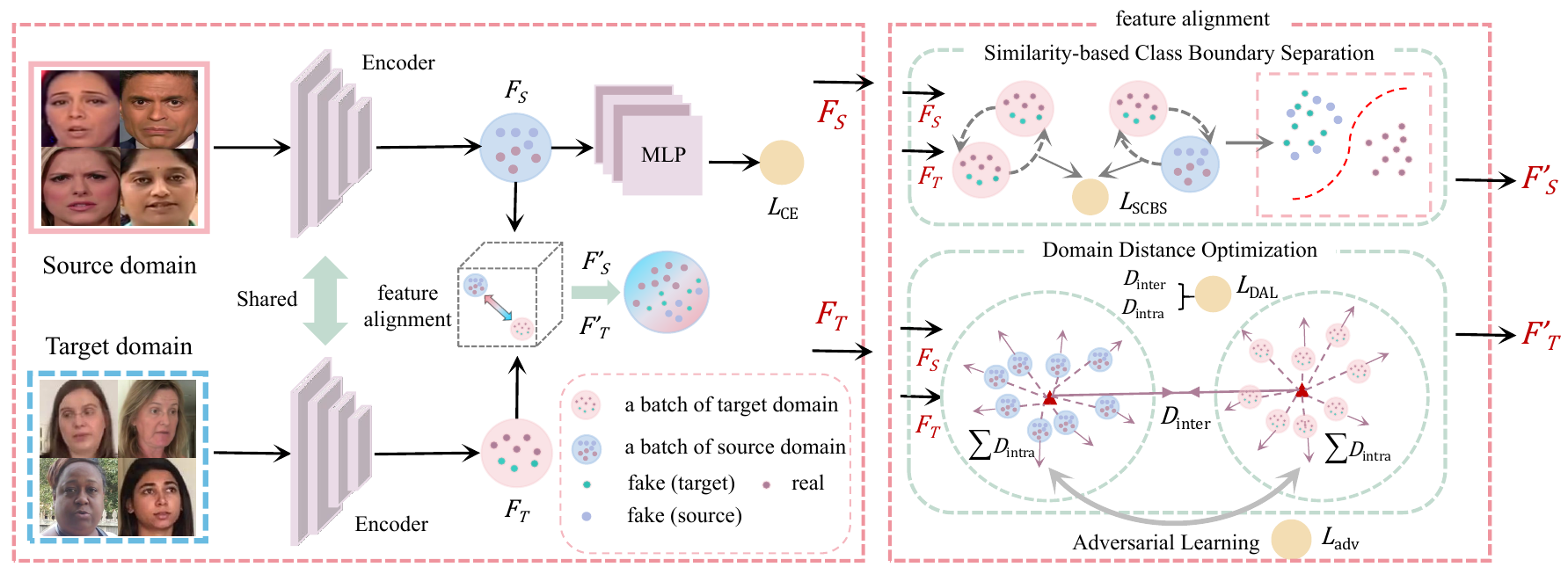} 
    \caption{The framework of the Open-World Generalization Enhancement Training Strategy (OWG-DS) includes the Domain Distance Optimization (DDO) module, the Similarity-Based Class Boundary Separation (SCBS) module, and the adversarial training mechanism. $F'_S$ and $F'_T$ are features that are fully aligned after training.} 
    \label{fig1:modeloverview} 
\end{figure*}

\subsection{Overview}
As shown in Figure~\ref{fig1:modeloverview}, a novel Open-World Deepfake Detection Generalization Enhancement Training Strategy (OWG-DS) is proposed. The proposed strategy is able to achieve robust feature representation learning by jointly optimizing different domain feature alignment and class boundary refinement, which effectively solves the domain shift problem between different domains in open world environment.

Specifically, the model $M$ is pre-trained on the labeled source domain $D_S$ to learn common forgery detection, laying the foundation for cross-domain feature learning. We then retrain the model jointly with source and target data, i.e., using labeled source data for supervised training and unlabeled target data for unsupervised adaptation. To this end, we first use the encoder $E$ of the pre-trained model $M$ to map $D_S$ and $D_T$ into a shared feature latent space, thereby obtaining source domain feature $F_S$ and target domain features $F_T$. $F_S$ and $F_T$ separately represent the feature distribution of the corresponding domain:
\begin{equation}
    F_S = \{f_i = E(x_i)\}_{i=1}^n
\end{equation}
\begin{equation}
    F_T = \{f_i = E(x_i)\}_{i=1}^m
\end{equation}
where $E(\cdot)$ denotes the feature extraction process of $M$.
Then, the Domain Distance Optimization (DDO) module is used to compute the global centroid $\mathbf{C}_{\text{global}}^{S}$ of $F_S$ and the global centroid $\mathbf{C}_{\text{global}}^{T}$ of $F_T$, respectively.
These centroids represent the central location of the corresponding domain feature distribution in the shared feature latent space, which can be used to quantify the inter-domain distance.
By reducing the distance between centroids, samples of different domains are guided to come closer together in feature space to achieve inter-domain convergence. Meanwhile, we further align different domain features by expanding the divergence between feature points within each domain.
The dual-objective optimization method maximizes the overlap between features from different domains, thereby mitigating domain shift.

Considering that feature space collapses and cross domain feature interference may occur during the above feature alignment process, leading to class boundary blurring and reducing category discriminability, a Similarity-based Class Boundary Separation (SCBS) module is proposed to mitigate these effects. The module uses a positive sample learning mechanism to model the similarity relationships among samples, identifying for each sample $f_i$ its most similar sample $f_j$ excluding itself to form positive pairs $(f_i, f_j)$. All positive pairs of samples are optimized to improve the similarity scores, ensuring that samples with similar characteristics (i.e., from the same class) are grouped closer in the feature space. This process reinforce the tight aggregation of similar samples while pushing samples from different classes apart, thereby enhancing the separation between category boundaries and minimizing the overlap between features in each category. In addition, to further mitigate domain shift, a adversarial optimization mechanism based on an adversarial domain classifier (ADC) is designed. This mechanism enhances cross-domain feature alignment through adversarial interaction between ADC and encoder $E$. 

Overall, the proposed OWG-DS realizes the alignment between different domain features through DDO and ADC, which is conducive to extracting cross-domain features with high generalization, while SCBS further ensures the inter-class differentiation ability of cross-domain features. The synergistic optimization of the three can greatly enhance the generalization ability of existing detection methods, effectively solving the deepfake detection generalization task in open scenarios. In the below, we detail the key modules in the following subsections.

\subsection{Domain Distance Optimization } \label{sec:DDOM}
Current domain alignment methods typically focus on reducing the distance between different domains to achieve feature alignment, thereby mitigating domain shift. However, merely reducing the inter-domain distance is not enough to fully align the features of different domains, and has a limited effect on the mitigation of domain shift. In order to maximize the feature alignment between two domains, we propose Domain Distance Optimization (DDO) module, which not only enhances the inter-domain compactness, but also expand the intra-domain divergence at the same time to achieve optimal domain shift reduction.

Specifically, we introduce global centroids $\mathbf{C_{\text{global}}}$ to calibrate the central locations of the different domain feature distributions in the shared latent space. Global centroids are constructed through a two-stage strategy: first, local centroids $\mathbf{C_{\text{local}}}$ are obtained by averaging the feature of small mini-batches. Then, the global centroids are iteratively refined based local centroids through a momentum-based update mechanism. Formally,
\begin{equation}
 \quad \mathbf{C}_{\text{local}} = \frac{1}{N} \sum_{i=1}^{N} {f}_i
  \label{eq:local_centroid}
\end{equation}

\begin{equation}
 \mathbf{C}_{\text{global}} \leftarrow \mu \cdot \mathbf{C}_{\text{global}} + (1 - \mu) \cdot \mathbf{C}_{\text{local}}
  \label{eq:global_centroid}
\end{equation}
where $N$ is the number of small-batch samples, $f_i$ represents the feature of the i-th sample in the domain, and $\mu$ denotes the momentum coefficient. It is noteworthy that global centroids are initialized to zero values. During the training process, the momentum-based update mechanism can effectively counteract the fluctuation of feature differences between batches and mitigate the impact of extreme features on the global centroid representation.

\begin{figure*}[t] 
    \centering
    \includegraphics[width=0.9\textwidth]{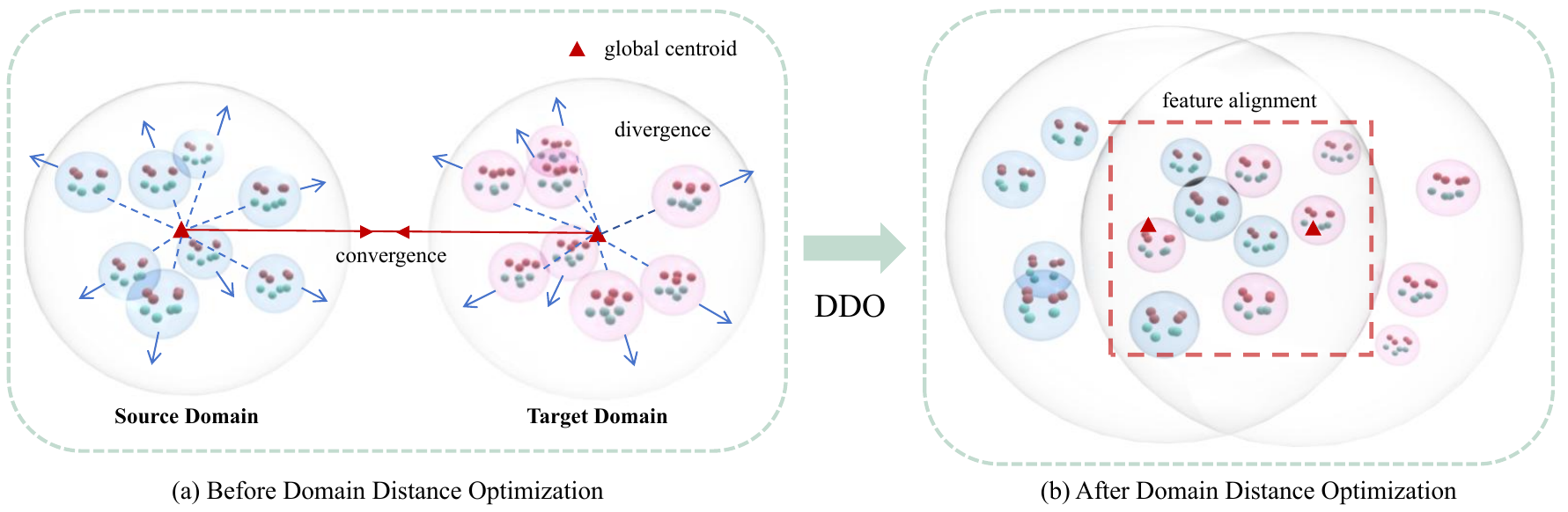} 
    \caption{The Domain Distance Optimization (DDO) diagram shows the dynamic process of cross-domain feature alignment by comparing the feature distribution before and after optimization. The left side (before optimization) displays the original distribution of the source domain (blue cluster) and target domain (pink cluster), with a noticeable shift between their centroids (red triangles). The right side (after optimization) shows the feature space after processing by DDO, where the centroids move closer together, and the samples spread out.} 
    \label{fig2:DDOM} 
\end{figure*}

To enhance the inter-domain compactness, we can obtain the global centroid of source domain $\mathbf{C}_{\text{global}}^{S}$ and the global centroid of target domain $\mathbf{C}_{\text{global}}^{T}$ according to Equation \ref{eq:local_centroid} and \ref{eq:global_centroid}, respectively. The Euclidean distance between these two centroids, denoted as $D_{\text{inter}}$, can be served as the metric to quantify the distance between different domains: 
\begin{equation}
 D_{\text{inter}} = \|\mathbf{C}_{\text{global}}^{S} - \mathbf{C}_{\text{global}}^{T}\|_2
  \label{eq:D_inter}
\end{equation}

The feature distribution of two domains can be brought closer by reducing the $D_{\text{inter}}$.
As illustrated in Figure \ref{fig2:DDOM} (a), we use spheres to model the distribution of source and target domain data in the feature space. The red triangles within the sphere represent the global centroids of the domain, and the distance between these two global centroids is the $D_{\text{inter}}$. By reducing this $D_{\text{inter}}$, the spheres are brought closer together, thereby increasing the compactness between the domains.

Meanwhile, we compute the intra-domain distance $D_{\text{intra}}$ (i.e., the average of the Euclidean distances of all sample features in domain to the global centroid) to characterize the size of domain space. Formally,
\begin{equation}
 D_{\text{intra}} = \frac{1}{N} \sum_{i=1}^{N} \|{f}_i - \mathbf{C}_{\text{global}}\|_2
  \label{eq:D_intra}
\end{equation}
where $N$ denotes the total number of samples in the domain, ${f}_i$ represent the feature of the i-th sample in the domain.

By obtaining the intra-domain distance for the source domain $D_{\text{intra}}^S$ and the target domain $D_{\text{intra}}^T$ according to Equation \ref{eq:D_intra}, the intra-domain divergence of source and target domains can be expanded by simultaneously increasing $D_{\text{intra}}^S$ and $D_{\text{intra}}^T$.  This process can be also illustrated in Figure \ref{fig2:DDOM}, by increasing the intra-domain distance, the volume of each sphere is gradually expanding to become larger, spreading the feature distribution within the sphere.

To simultaneously increase the inter-domain compactness and expand the intra-domain divergence, we introduce a domain alignment loss: 
\begin{equation}
 \mathcal{L}_{\text{DAL}} = D_{\text{inter}} + \exp(-(D_{\text{intra}}^S+ D_{\text{intra}}^T)) \cdot w_{\text{intra}}
  \label{eq:L_DAL}
\end{equation}
where \( w_{\text{intra}} = 1 - \frac{\text{epoch}}{\text{Epoch}} \). It is a dynamic weight to control the rate of intra-domain  divergence. "\text{Epoch}" represents the total number of training epochs, and "\text{epoch}" refers to the current training epoch. Through this dynamic adjustment, a more balanced intra-domain feature distribution can be achieved at different training stages. The dual-objective optimization of the domain alignment loss maximizes the overlap between features from different domains, thereby effectively mitigating domain shift.

\subsection{Similarity-based Class Boundary Separation}
In Section \ref{sec:DDOM}, we optimize the distance between the two domains to align different domain features, thereby mitigating domain shift. However, feature space collapses and cross domain feature interference may occur during the above feature alignment, leading to class boundary blurring and reducing category discriminability of domain features. To address this issue, we propose a Similarity-based Class Boundary Separation (SCBS) module. 

The SCBS module uses a positive sample learning mechanism to model the similarity relationships among samples, identifying for each sample $f_i$ its most similar sample $f_j$ excluding itself to form positive pairs $(f_i,  f_j)$. Specifically, for labeled source domain samples, we build cross sample positive pairs based on label semantics by randomly selecting non-identical samples from the same class as positive examples. For unlabeled target domain samples, we adopt a feature space cosine similarity-based neighbor screening method, where the second nearest neighbor is selected as pseudo-positive pairs. While cosine similarity-based neighbor screening method may carry the risk of erroneously pairing samples from different classes, the initial discriminative ability of the pre-trained feature extractor makes it more likely that correct pseudo-positive pairs will dominate in the target domain, thereby significantly reducing this risk.
The core idea behind this positive sample learning mechanism is to encourage samples belonging to the same class (or inferred as such based on feature similarity) to cluster closer together in the feature space by maximizing the similarity of the constructed positive pairs $(f_i,  f_j)$. This process compels the model to learn representations where similar/same-class samples are grouped together. Consequently, this effectively increases intra-class compactness and enlarges inter-class distances, leading to clearer class boundaries.

To enforce tighter aggregation of same class samples and sharpen different classes boundaries, we design the $\mathcal{L}_{\text{SCBS}}$ to optimize feature similarity for positive pairs. The loss is designed to encourage compact intra-class sample aggregation in the feature space while maintaining clear separation between different classes. 

This $\mathcal{L}_{\text{SCBS}}$ can be defined as
\begin{equation}
\mathcal{L}_{\text{SCBS}} = -\frac{1}{|\mathcal{P}|} \sum_{(i,j) \in \mathcal{P}} \log \sigma(s_{ij})
\end{equation}
where $|\mathcal{P}|$ denotes the total number of positive samples in the $\mathcal{P}$, $s_{ij}$ denotes the cosine similarity between the features $f_i$ and $f_j$ and $\sigma$ denotes the softmax function.

\subsection{Adversarial Domain Classification}
To further mitigate domain shift, we employ an adversarial optimization mechanism based on an adversarial domain classifier (ADC). The mechanism adopts a dynamic gradient modulation to establish an adversarial relationship between the domain classifier and feature extraction, thereby enhancing cross-domain feature alignment. This adversarial optimization method enables feature representations to gradually free themselves from domain-related interference factors, forming a general feature pattern with cross-domain adaptability.
During training, we use the binary cross-entropy loss $\mathcal{L}_{\text{adv}}$ to supervise the optimization of the domain classifier. 

Then the $\mathcal{L}_{\text{adv}}$ can be denoted as
\begin{equation}
  \small
  \mathcal{L}_{\text{adv}} = - \frac{1}{N_S + N_T} \sum_{i=1}^{N_S + N_T} \left[ y_i \log(\hat{y}_i) + (1 - y_i) \log(1 - \hat{y}_i) \right]
  \label{eq:L_domain}
\end{equation}
where $N_S$ and $N_T$ respectively represent the number of samples in the source domain and target domain, while $y_i$ and $\hat{y}_i$ respectively represent the true and predicted label of domain classification.

\subsection{Loss Functions}
In addition to the constraints mentioned above, we also use a regularization term $R$ \cite{cao2021open} during training to avoid simple solutions that assign all instances to the same class. The formula is as follows:
\begin{equation}
  R = \text{KL} \left( \frac{1}{N_S + N_T} \sum_{x_i \in D_S \cup D_T} \sigma(F(x_i)) \| P(y) \right)
  \label{eq:R}
\end{equation}
where $p_i = \sigma(F(x_i))$ is class prediction probability and $P$ denotes a prior probability distribution of label $y$.

We introduce KullbackLeiler (KL) divergence regularization to constrain the class distribution of the model output, ensuring that it is closer to the prior distribution $P$. Since assuming knowledge of the prior distribution is often unrealistic, we apply maximum entropy regularization to avoid relying on such assumptions. The use of regularization keeps the output of the model somewhat diverse, thus avoiding an overly flat distribution.

For the labeled source domain, we employ traditional supervised training, using the cross-entropy loss $L_{CE}$ to compute the loss.

The final loss function is given by:
\begin{equation}
 \mathcal L = \mathcal L_{\text{CE}} + \eta_1 \mathcal L_{\text{DAL}} + \eta_2 \mathcal L_{\text{SCBS}} + \eta_3 \mathcal L_{\text{adv}} +\eta_4 R,
  \label{eq:L}
\end{equation}
with hyper-parameters $\eta_1$, $\eta_2$, $\eta_3$ and $\eta_4$.



\begin{table*}
  \caption{Cross-manipulation detection results using FF++ HQ/LQ data are reported, with three forgery methods as source domains and the remaining method as the target, compared with state-of-the-art domain adaptation methods.}
  \label{tab:cross-methods}

  \begin{tabular}{c|p{0.62cm}p{0.62cm}p{0.62cm}p{0.62cm}|p{0.62cm}p{0.62cm}p{0.62cm}p{0.62cm}|p{0.62cm}p{0.62cm}p{0.62cm}p{0.62cm}|p{0.62cm}p{0.62cm}p{0.62cm}p{0.62cm}}
    \toprule
   \multirow{3}{*}{\textbf{Methods}} & \multicolumn{4}{c|}{\textbf{DF FS NT $\rightarrow$ FF}} & \multicolumn{4}{c|}{\textbf{FF FS NT $\rightarrow$ DF}} & \multicolumn{4}{c|}{\textbf{DF FF FS $\rightarrow$ NT}} & \multicolumn{4}{c}{\textbf{DF FF NT $\rightarrow$ FS}} \\ \cline{2-17}

   & \multicolumn{2}{c|}{\textbf{source}} &\multicolumn{2}{c|}{\textbf{target}} &\multicolumn{2}{c|}{\textbf{source}} & \multicolumn{2}{c|}{\textbf{target}} & \multicolumn{2}{c|}{\textbf{source}} &\multicolumn{2}{c|}{\textbf{target}} & \multicolumn{2}{c|}{\textbf{source}} &\multicolumn{2}{c}{\textbf{target}}\\
    \cline{2-17}

    & \textbf{Acc} & \textbf{Auc} & \textbf{Acc} & \textbf{Auc} & \textbf{Acc} & \textbf{Auc} & \textbf{Acc} & \textbf{Auc} & \textbf{Acc} & \textbf{Auc} & \textbf{Acc} & \textbf{Auc} & \textbf{Acc} & \textbf{Auc} & \textbf{Acc} & \textbf{Auc} \\
    \hline

    \multicolumn{17}{c}{\textbf{LQ (Low Quality)}} \\ \hline

    {\textbf{Xception}} & 76.26 & 84.85 & 58.29 & 64.08 & 74.81 & 82.98 & 57.52 & 68.65 & 83.35 & 92.12 & 57.44 & 60.65 & 78.62 & 87.97 & 56.25 & 60.04 \\ \hline 
                                   
    {\textbf{ORCA}} & 75.17 & 86.34 & 62.00 & 66.24 & 76.00 & 84.31 & 79.90 & 90.31 & 80.12 & \underline{94.29} & 59.92 & \underline{61.03} & 79.03 & 89.47 & 58.17 & 59.53 \\ \hline 

    {\textbf{OSDD}} & \textbf{82.04} & \textbf{90.93} & 65.36 & 71.99 & \textbf{78.20} & \textbf{87.11} & 79.13 & 86.06 & \textbf{90.08} & \textbf{96.72} & \textbf{61.21} & \textbf{63.33} & \textbf{84.42} & \textbf{92.04} & 64.11 & 67.44 \\ \hline 

    {\textbf{Ours}} & \underline{78.48} & \underline{88.69} & \textbf{67.40} & \textbf{73.37} & \underline{77.95} & \underline{86.74} & \textbf{81.98} & \textbf{90.45} & \underline{87.69} & 84.99 & \underline{60.24} & 60.67 & \underline{82.34} & \underline{91.05} & \textbf{64.35} & \textbf{68.32} \\ \hline

    \multicolumn{17}{c}{\textbf{HQ (High Quality)}} \\ \hline
    
    {\textbf{Xception}} & 92.30 & 98.01 & 68.59 & 76.44 & 90.68 & 96.96 & 67.24 & 74.63 & 96.61 & 99.50 & 63.71 & 71.75 & 94.91 & 98.89 & 57.20 & 54.58 \\ \hline 

    {\textbf{DF}} & 89.65 & 97.45 & 76.79 & 83.83 & 89.54 & 96.19 & 82.54 & 90.25 & 92.35 & 98.53 & 69.25 & 74.89 & 91.26 & 98.19 & 63.79 & 71.85 \\ \hline

    {\textbf{FreqNet}} & 80.27 & 92.68 & 58.08 & 60.11 & 82.23 & 94.98 & 63.06 & 75.97 & 83.18 & 91.74 & 65.79 & 74.10 & 88.01 & 95.14 & 55.02 & 54.64 \\ \hline

    {\textbf{OSDD}} & \underline{93.29} & \underline{98.66} & 64.15 & 79.89 & \underline{89.77} & \underline{98.01} & 76.27 & 88.62 & \underline{97.53} & \underline{99.66} & 61.83 & 75.75 & \underline{93.73} & \underline{98.30} & \underline{69.88} & \underline{73.56} \\ \hline

    {\textbf{ORCA}} & 90.46 & 97.30 & \underline{83.33} & \underline{91.53} & 87.74 & 95.33 & \underline{82.58} & \underline{92.17} & 86.68 & 98.11 & \underline{70.81} & \underline{78.04} & \textbf{94.63} & \textbf{99.13} & 56.84 & 52.12 \\ \hline 

    {\textbf{Ours}} & \textbf{94.22} & \textbf{98.94} & \textbf{89.92} & \textbf{97.31} & \textbf{93.13} & \textbf{97.96} & \textbf{89.76} & \textbf{96.14} & \textbf{97.99} & \textbf{99.72} & \textbf{78.15} & \textbf{90.02} & 92.90 & 98.24 & \textbf{91.23}\textbf & \textbf{96.92} \\
  \bottomrule
\end{tabular}
\end{table*}

\section{Experiments}

\subsection{Experimental Setting}

\textbf{Datasets.} To validate the proposed method, we conducted extensive experiments on three datasets: FF++ \cite{rossler2019faceforensics++}, Celeb-DF \cite{li2020celeb}, and DFDC \cite{dolhansky2020deepfake}. The FF++ dataset includes four manipulation techniques: Deepfake (DF), Face2Face (FF), FaceSwap (FS), and NeuralTextures (NT), with three video versions: original, high-quality (HQ), and low-quality (LQ). The Celeb-DF dataset captures real-world challenges such as pose variations, lighting, and facial expressions, making it suitable for diverse deepfake detection tasks. The DFDC dataset uses multiple forgery techniques, including Deepfake, FaceSwap, and FSGAN, simulating real-world forgery methods. In our experiment, we focus on frame-level deepfake detection, randomly selecting frames from the videos in the dataset for training. Additionally, the same face cropping method is applied to the video frames in all three datasets.

\textbf{Domain adaptive scenario.}
To validate our proposed domain adaptation method, we designed two scenarios: Cross-manipulation-methods and Cross-manipulation-datasets. Cross-manipulation-methods assesses the model's ability to transfer across different manipulation methods within the FF++ dataset (e.g., from Deepfake to Face2Face). Cross-manipulation-datasets evaluates the model's ability to transfer between datasets (e.g., from FF++ to Celeb-DF). In each scenario, the source domain is the labeled dataset, and the target domain is the unlabeled one. We train the model using both source and target domain data, then evaluate it on both domains. To simulate open world conditions, we ensure the target domain has a larger sample size than the source domain, validating the model's ability to generalize with large-scale unlabeled data.

\textbf{Implementation Details.}
We pretrain the Xception \cite{chollet2017xception} model on the source domain dataset for 150 epochs with a $224 \times 224$ input image size to equip the model with an initial ability to detect common Deepfake data. After pretraining, we apply the proposed training strategy for an additional 100 epochs. The loss term weights are set as $\eta_1 = 0.1$, $\eta_2 = 1$, $\eta_3 = 1$, and $\eta_4 = -1$. The batch size for training data is computed based on the ratio of labeled to unlabeled data, ensuring that the total batch size for both types of data is 50, while the batch size for testing data is set to 32. We use stochastic gradient descent (SGD) as the optimizer, with a learning rate of 0.0005, momentum of 0.9, and weight decay of 0.0005. All experiments are conducted on an NVIDIA GeForce RTX 3090 GPU using the Pytorch framework.

\begin{figure*}[h] 
\centering
\includegraphics[width=\textwidth]{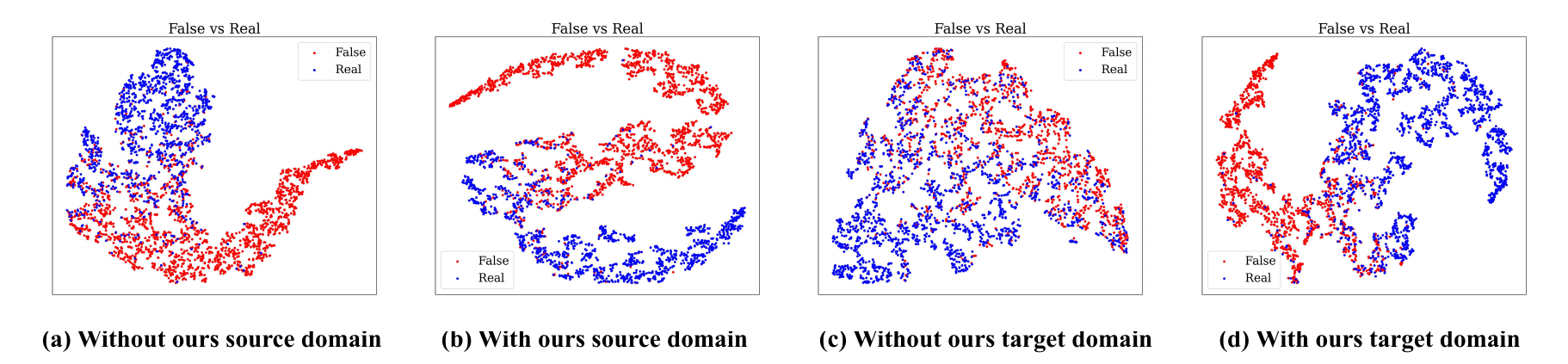} 
\caption{T-SNE visualization on FF++ (FF FS NT $\rightarrow$ DF) (HQ).} 
\label{fig:T-SNE} 
\end{figure*}

\subsection{Cross manipulation methods}
We designed an experiment using four forgery methods from the FF++ dataset (DF, FF, NT, FS), selecting any three methods as the source domain and a method different from the three source domain methods as the target domain, to evaluate the performance of our method across different forgery methods.

The performance results in the cross-manipulation-methods scenario are shown in Table \ref{tab:cross-methods}, with high-quality (HQ) data in the lower half and low-quality (LQ) data in the upper half. To more clearly demonstrate the superiority of our method over previous domain adaptation approaches, all methods listed in the table use the Xception model as the backbone network. The only exception is FreqNet \cite{tan2024frequency}, which incorporates an additional frequency feature extraction module, making its backbone different from Xception.

Experimental results demonstrate that our method surpasses existing approaches in both the source and target domains, exhibiting stronger domain adaptation capabilities. In the HQ setting, our method achieves a highest source domain ACC of 97.99\% and a highest target domain ACC of 89.92\%, with a maximum improvement of 34.03\% compared to the baseline. Furthermore, compared to existing domain adaptation methods such as ORCA \cite{cao2021open} and DomainForensics (abbreviated as DF in the table) \cite{lv2024domainforensics}, our method preserves source domain knowledge during adaptation, preventing catastrophic forgetting and ensuring that target domain adaptation does not degrade the detection capability of the source domain. Notably, in the DF FF NT $\rightarrow$ FS task, the performance of target domain is constrained across all baseline methods, likely due to the significant distribution gap between the source and target domains, which increases the difficulty of domain adaptation. However, even under this challenging setting, our method still achieves competitive performance, further validating its generalization ability in complex cross-domain tasks.

The Figure \ref{fig:T-SNE} presents the T-SNE feature distribution visualization for (FF FS NT $\rightarrow$ DF). It shows the following four scenarios: source and target domain feature distributions under both with/without our method configurations. For the target domain, the visualization clearly shows that, compared to the mixed feature distribution without using our method, the features of real and fake data are better separated. Notably, in the source domain's mixed feature distribution, our method enhances categories separation compared to its absence. This indicates that our proposed training strategy effectively adapts to the new domain's feature distribution while preserving the recognition capability of the source domain's features, without losing the memory of the original domain's features.

\subsection{Cross manipulation datasets}
In the cross-manipulation-datasets scenario, we use the four forgery methods (DF, FF, FS, NT) from the FF++ (HQ)  dataset as the source domain, and the Celeb-DF and DFDC datasets as the target domains. The performance results are shown in Table \ref{tab:cross-datasets}. Consistent with the setup in Section 4.2, Xception serves as the "baseline" method, meaning it is trained on the source domain and directly tested on the target domain. To ensure a fair comparison, all domain adaptation methods listed in the table adopt Xception as their backbone network. The reported values represent AUC scores.

Experimental results demonstrate that our method achieves superior performance in both the source and target domains. In the FF++ $\rightarrow$ Celeb-DF task, our method achieves an AUC of 99.51\% on the target domain, representing an improvement of 27.18\% over Xception \cite{chollet2017xception} and 19.30\% over the latest competing method, OSDD \cite{zhou2024fine}, highlighting the stronger adaptability of our method to unseen forgery data. Similarly, in the FF++ $\rightarrow$ DFDC task, our method attains an AUC of 89.37\%, outperforming Xception by 23.76\% and ORCA \cite{cao2021open} by 8.56\%. Notably, while our method exhibits slightly lower performance than OSDD in the source domain, it surpasses OSDD by 19.30\% and 16.70\% on the Celeb-DF and DFDC target datasets, respectively. This demonstrates that our method achieves a better balance between preserving source domain knowledge and improving detection performance of the
target domain. Our method effectively maintains the recognition ability of the source domain while significantly enhancing adaptability to the unknown target domain, thereby validating its generalization advantage in cross-domain scenarios.






\subsection{Ablation Study}
\subsubsection{Effectiveness of different components: }

We first investigated the indispensability of each component in OWG-DS. Specifically, we removed the Domain Distance Optimization (DDO) module, the Similarity-Based Class Boundary Separation (SCBS) module, and the Adversarial Domain Classification (ADC) module to evaluate the contribution of each component to the overall architecture. The experimental results on the FF++ (DF FS NT → FF) (HQ) dataset are shown in Table \ref{tab:Ablation}. The results indicate that the performance drops whenever a module is removed. Notably, the SCBS module has the greatest impact on performance, as it adaptively adjusts the class boundaries between real and fake classes. This works in synergy with DDO to effectively prevent overlap between different classes during the domain feature alignment process in DDO. Overall, each component of OWG-DS plays a crucial role and has been carefully designed.

\begin{table}
  \caption{In the cross-dataset scenarios from FF++ to Celeb-DF and from FF++ to DFDC, we compare the performance of our method with the state-of-the-art methods.}
  \label{tab:cross-datasets}
  \begin{tabular}{c|cc|cc}
    \toprule
   \multirow{2}{*}{\textbf{Methods}} & \multicolumn{2}{c|}{\textbf{FF++ $\rightarrow$ Celeb-DF}} & \multicolumn{2}{c}{\textbf{FF++ $\rightarrow$ DFDC}}\\ \cline{2-5}

   & \textbf{source} & \textbf{target} & \textbf{source} & \textbf{target} \\
    \hline
    
    {\textbf{Xception}} & 95.62  & 72.33   & 95.62 & 65.61 \\ \hline

    {\textbf{ORCA}} & 91.01  & \underline{97.08}   & 92.14  & \underline{80.81}  \\ \hline

    {\textbf{DomainForensics}} & 86.68  & 89.02   & 91.05  & 80.67  \\ \hline 
    
    {\textbf{OSDD}} & \textbf{96.79}  & 80.21   & \textbf{96.44}  & 72.67  \\ \hline 

    {\textbf{Ours}} & \underline{95.85}  & \textbf{99.51}   & \underline{95.78}  & \textbf{89.37} \\
  \bottomrule
\end{tabular}
\end{table}

\begin{table}[t]
\centering
\caption{Ablation study with different components.}
\begin{tabular}{c c c | c c}
\hline
\multirow{2}{*}{\textbf {DDO}} & \multirow{2}{*}{\textbf{SCBS}} & \multirow{2}{*}{\textbf{ADC}} & \multirow{2}{*}{\textbf {Acc}} & \multirow{2}{*}{\textbf{$\Delta$}} \\
 & & & & \\
\hline
& \checkmark &\checkmark & 86.15 & -3.77 \\
\hline
\checkmark & & \checkmark & 77.60 & -12.32\\
\hline

\checkmark & \checkmark & & 84.11 & -5.81\\
\hline

\checkmark & \checkmark & \checkmark & 89.92 & 0.00 \\
\hline
\label{tab:Ablation}
\end{tabular}
\end{table}

\subsubsection{Data efficiency of the proposed method: }
\begin{figure}[t] 
\centering
\includegraphics[width=1.0\linewidth]{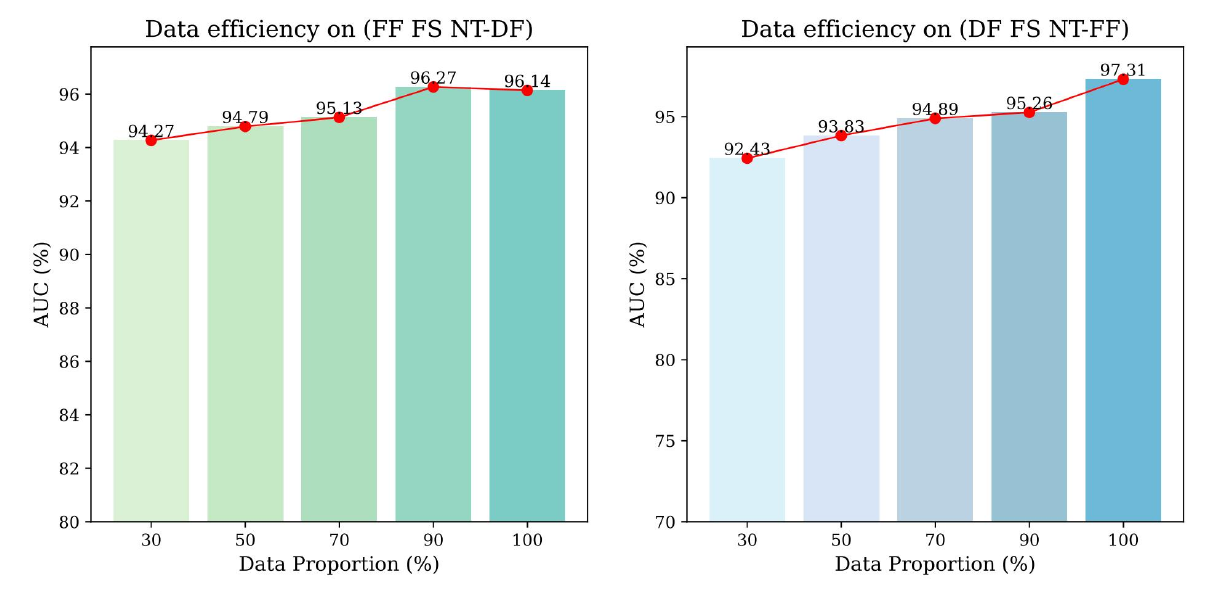} 
\caption{Data efficiency on (DF FS NT-FF) and (FF FS NT-DF)} 
\label{fig:data_proportion} 
\end{figure}

To simulate open world data constraints, we randomly selected different proportions of target domain samples to evaluate the data efficiency of the proposed method. This experiment was conducted on the FF++ (HQ) dataset in the (DF FS NT $\rightarrow$ FF) and (FF FS NT $\rightarrow$ DF) scenarios. Figure \ref{fig:data_proportion} presents the results with target domain data proportions of 30\%, 50\%, 70\%, 90\%, and 100\%. While reducing the target domain data leads to a performance drop, using only 30\% of the target data results in a decrease of just 4.88\% and 1.87\% compared to using 100\%. Notably, even with only 30\% of the target domain data, our method still outperforms all methods in Table \ref{tab:cross-methods}, surpassing Xception by 15.99\% and 19.64\% in the two scenarios, respectively. This demonstrates that our method remains highly adaptable even under limited data conditions.

\subsubsection{Different models versatility of the proposed method: }

Our proposed method is independent of specific model architectures. Table \ref{tab:models versatility} presents the performance of applying our method to Xception, ResNet-50 \cite{he2016deep}, and EfficientNet-B0 \cite{tan2019efficientnet} on the FF++ (HQ) dataset. The reported values represent AUC scores. The results demonstrate that regardless of the model architecture, our method significantly improves performance and enhances generalization to the target domain. Moreover, during adaptation from the source domain to the target domain, our method effectively preserves source domain knowledge, preventing forgetting effects. 

\begin{table}[t]
  \caption{For FF++ (HQ), we use three forgery methods as the source domain and adapt to the fourth. We compare Xception, ResNet-50, and EfficientNet-B0 with and without our method.}
  \label{tab:models versatility}
  \begin{tabular}{c|cc|cc}
    \toprule
   \multirow{2}{*}{\textbf{Methods}} & \multicolumn{2}{c|}{\textbf{DF FS NT $\rightarrow$ FF}} & \multicolumn{2}{c}{\textbf{FF FS NT $\rightarrow$ DF}}\\ \cline{2-5}

   & \textbf{source} & \textbf{target} & \textbf{source} & \textbf{target} \\
    \hline
    
    {\textbf{Xception}} & 98.01  & 76.44  & 96.96 & 74.63 \\ \hline

    {\textbf{Xception + Ours}} & 98.94  & 97.31   & 96.96  & 96.14  \\ \hline    \hline

    {\textbf{ResNet-50}} & 96.60  & 80.79  & 79.29  & 71.04  \\ \hline 

    {\textbf{ResNet-50 + Ours}} & 98.16  & 94.28  & 85,77  & 85.35  \\ \hline    \hline
    
    {\textbf{EfficientNet-B0}} & 96.40  & 78.47  & 95.36  & 75.96  \\ \hline 

    {\textbf{EfficientNet-B0 + Ours}} & 98.47  & 92.60  & 97.37  & 96.25  \\ \hline    \hline

    \multirow{2}{*}{\textbf{Methods}} & \multicolumn{2}{c|}{\textbf{DF FF FS $\rightarrow$ NT}} & \multicolumn{2}{c}{\textbf{DF FF NT $\rightarrow$ FS}}\\ \cline{2-5}

   & \textbf{source} & \textbf{target} & \textbf{source} & \textbf{target} \\
    \hline
    
    {\textbf{Xception}} & 99.50  & 71.75  & 98.89 & 54.58 \\ \hline

    {\textbf{Xception + Ours}} & 99.72  & 90.14  & 98.24  & 96.92  \\ \hline    \hline

    {\textbf{ResNet-50}} & 96.57  & 76.68  & 86.64  & 58.03  \\ \hline 

    {\textbf{ResNet-50 + Ours}} & 98.76  & 86.66  & 92.46  & 90.31  \\ \hline    \hline
    
    {\textbf{EfficientNet-B0}} & 98.28  & 64.47   & 97.48  & 51.28  \\ \hline 

    {\textbf{EfficientNet-B0 + Ours}} & 98.95  & 73.84  & 98.96  & 90.10  \\ 
    
  \bottomrule
\end{tabular}
\end{table}

\section{Conclusions}
In this paper, we define a deepfake detection generalization task which focuses on how to achieve efficient detection of large amounts of unlabeled data based on limited labeled data to simulate a open-world scenario. To solve this task, we propose a novel open-world training strategy termed OWG-DS to improve the generalization ability. Specifically, we introduce the DDO module, which optimizes inter-domain and intra-domain distances to align different domain features and mitigate domain shifts. Additionally, we introduce the SCBS module to enhance the aggregation of similar samples, thereby ensuring clearer class boundaries. Furthermore, we adopt an adversarial training mechanism to guide the feature extractor in learning domain-invariant features, enhancing cross-domain generalization. Extensive experiments show that our method excels in cross-method and
cross-dataset scenarios while maintaining strong detection performance even with limited target domain data. Moreover, we validate the model-agnostic nature of our approach, showing that it can be seamlessly integrated into various architectures to significantly improve their generalization capability.


\bibliography{main}

\appendix

\end{document}